\documentclass[10pt,twocolumn,letterpaper]{article}

\usepackage{subfig}

\usepackage{iccv}
\usepackage{times}
\usepackage{epsfig}
\usepackage{graphicx}
\usepackage{amsmath}
\usepackage{amssymb}
\usepackage{rotating}

\usepackage[pagebackref=true,breaklinks=true,letterpaper=true,colorlinks,bookmarks=false]{hyperref}

\iccvfinalcopy 

\def\iccvPaperID{956} 

\ificcvfinal\fi
\begin{document}

\title{Semantic Instance Segmentation via Deep Metric Learning}

\author{Alireza Fathi\thanks{Google Inc., USA} \and Zbigniew Wojna\footnotemark[1] \and Vivek Rathod\footnotemark[1] \and Peng Wang\thanks{UCLA} \and Hyun Oh Song\footnotemark[1] \and Sergio Guadarrama\footnotemark[1] \and Kevin P. Murphy\footnotemark[1]}

\maketitle

\newcommand{\eat}[1]{}
\newcommand{\todo}[1]{{\bf TODO: #1}}
\newcommand{\iouthresh}{\beta}
\newcommand{\APr}{$\mathrm{AP}^{r}$}
\newcommand{\mAPr}{m\APr}
\newcommand{\iou}{IoU}
\newcommand{\IOU}{\iou}
\newcommand{\topmap}{62.21 \%} 

\begin{abstract}
  We propose a new method for semantic instance segmentation,
by  first computing how likely two pixels are to belong to the same
object, and then by grouping similar pixels together.
Our similarity metric is based on a deep,
fully convolutional embedding model.
Our grouping method is based on
selecting all points that are sufficiently similar
to a set of ``seed points',
chosen from
 a deep, fully convolutional scoring model.
We show competitive results on
the Pascal VOC instance segmentation benchmark.
\eat{
  We propose a new method for semantic instance segmentation based on
  three key ideas. First, we can learn a measure of how likely two
  pixels are to belong to the same object using a deep embedding
  model. Second, we can learn a measure of how likely a pixel 
is to generate a mask that strongly overlaps with an object
instance. Third, we can combine these two models to find object
instances as follows: first, we identify good ``seed'' points, which
correspond to object centers; then we grow these seeds into entire
regions, based on identifying all the nearby pixels that are likely to
belong to the same instance. Finally, We show competitive results on
the Pascal VOC instance segmentation benchmark.
}
\end{abstract}

\section{Introduction}

Semantic instance segmentation is the problem of identifying
individual instances of objects and their
categories (such as person and car) in an image.
It differs from object detection in that the output is a mask
representing the shape of each object, rather than just a bounding
box.
It differs from semantic segmentation in that
our goal is not just to classify each pixel
with a label (or as background),
but also to
distinguish individual instances of the
same class. Thus, the label space is unbounded in size
(e.g., we may have the labels ``person-1'', ``person-2'' and
``car-1'', assuming there are two people and one car).
This problem has many practical applications in domains such as
self-driving cars, robotics, photo editing, etc. 

A common approach to this problem
(e.g., \cite{Hariharan2014,chen2015multi,Liang2016,Dai2016mnc})
is first to use some mechanism to predict object bounding boxes
(e.g., by running a class-level object detector, or by using a class agnostic
box proposal method such as EdgeBoxes),
and then to run segmentation and classification within each proposed
box.
However, this can fail if there is more than one instance inside of
the box.
Also, intuitively it feels more ``natural'' to first detect the
mask representing each object, and then derive a bounding box from
this, if needed.
(Note that boxes  are a good approximation to the shape of certain
classes, such as cars and pedestrians, but they are a poor approximation
for many other classes, such as articulated people,
``wirey'' objects like chairs, or non-axis-aligned objects like ships
seen from the air.)

\begin{figure*}[t]
\begin{centering}
\includegraphics[scale=0.3]{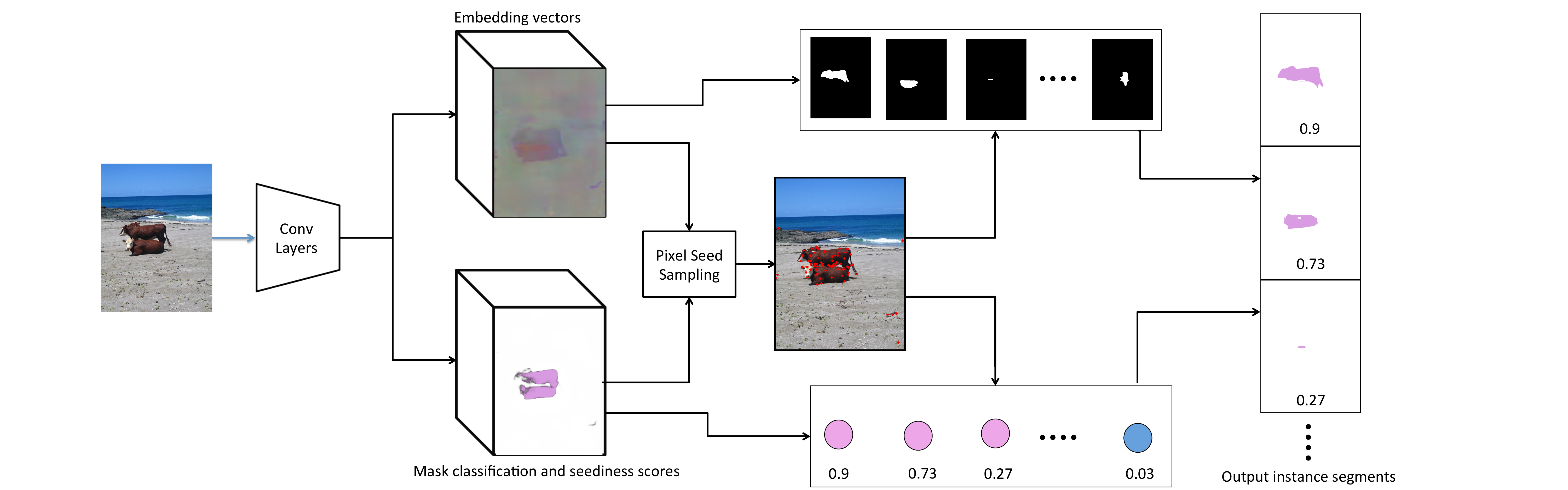}
\par
\end{centering}
\caption{Given an image, our model predicts the embedding vector of each pixel (top head of the network) as well as the classification score of the mask each pixel will generate if picked as a seed (bottom head of the network). We derive the seediness scores from the classification scores and use them to choose which seed points in the image to sample. Each seed point generates a mask based on the embedding vectors; each mask is then associated with a class label and confidence score. In this figure, pink color corresponds to the ``cow'' class and blue to the ``background'' class.
\label{fig:Model}
}
\end{figure*}

Recently there has been a move towards ``box-free'' methods, that try
to directly predict the mask for each object
(e.g., \cite{Pinheiro2015,Pinheiro2016,Hu2016,liang2015proposal}).
The most common approach to this is to modify the Faster RCNN
architecture \cite{Ren2016pami} so that at each point, it predicts
a ``centeredness'' score (the probability the current pixel is the
center of an object instance), 
a binary object mask,
and a class label
(rather than predicting the usual ``objectness'' score,
bounding box, and class label).
However, this approach requires that the entire object instance fits
within the receptive field of the unit that is making the
prediction. This can be difficult for elongated structures, that might
span many pixels in the image.
In addition, for some object categories, the notion of a ``center'' is
not well defined.

In this paper, we take a different approach.
Our key idea is that we can produce instance segmentations by
computing the likelihood that two pixels belong to the same object
instance and subsequently use these likelihoods to group similar
pixels together.
This is similar to most approaches
to  unsupervised image segmentation
(e.g. \cite{Shi2000,Felzenszwalb2004}),
which group pixels together to form segments or ``super-pixels''.
However, unlike the unsupervised case,
we have a well-defined notion of what a ``correct'' segment is, namely the
spatial extent of the entire object.
This avoids ambiguities such as whether to treat
parts of an object (e.g., the shirt and pants of a person)
as separate segments,
which plagues evaluation of unsupervised  methods.

We propose to learn the  similarity metric
using a deep embedding model.
This is similar to other approaches,
such as FaceNet \cite{Schroff2015},
which learn how likely two bounding boxes are to belong to the same
instance (person),
except we learn to predict the similarity of
{\em pixels},
taking into account their local context.

Another difference from unsupervised image segmentation is that we do
not use spectral or graph based partitioning methods,
because computing the pairwise similarity of all the pixels is too expensive.
Instead, we use compute the distance (in embedding space) to a set of
$K$ ``seed points''; this can be implemented as tensor multiplication.
To find these seed points, we learn a separate model that predicts
how likely a pixel is to make a good seed;
we call this the ``seediness'' score of each pixel.
This is analogous to the ``centeredness'' score used in prior methods, 
except we do not need to identify object centers; instead,
the seediness score is a measure of the ``typicality'' of a
pixel with respect to the other pixels in this instance.
We show that in practice we only have to take the top 100 seeds to
obtain good coverage of nearly all of the objects in an image.

Our method obtains a mAP score (at an IoU of 0.5)
of \topmap\ on the Pascal VOC 2012
instance segmentation benchmark \cite{Everingham2014}.
This puts us in fourth place, behind
\cite{Liang2016} (66.7\%),
\cite{li2016fully} (65.7\%),
and \cite{Dai2016mnc} (63.5\%).
However, these are all proposal-based methods.
(The previous fourth place
method was the ``proposal free'' approach of \cite{liang2015proposal},
who obtained 58.7\%.)
Although not state of the art on this particular benchmark,
our results are still competitive.
Furthermore, we believe our approach may work particularly well on
other datasets, with ``spindly'' objects with unusual shapes,
although we leave this to future work.

\section{Related work}

The most common approach to instance segmentation is first to predict
object bounding boxes, and then to predict the mask inside of each
such box using one or more steps.
For example, the MNC method of \cite{Dai2016mnc}, which won the COCO
2015 instance segmentation competition, was based on this approach. In
particular, they modified the Faster RCNN method \cite{Ren2016pami} as
follows: after detecting the top $K$ boxes, and predicting their class
labels, they extract features for each box using ROI pooling applied
to the feature map and use this to predict a binary mask representing
the shape of the corresponding object instance. The locations of the
boxes can be refined, based on the predicted mask, and the process repeated.
\cite{Dai2016mnc}
iterated this process twice, and 
the R2-IOS method in \cite{Liang2016} iterated the process a variable number of times.

\eat{
The R2-IOS method in \cite{Liang2016}
is somewhat similar to MNC,
except they use selective search to create the initial set of boxes,
rather than the Region Proposal Network (RPN) as in Faster RCNN. Also,
they use a variable number of iterations of the refinement process,
depending on the instance.  The R2-IOS currently has the best results
on the Pascal VOC 2012 validation set (66.7\% \mAPr\ at 0.5 IoU).
}

The fully convolutional instance segmentation
method of \cite{li2016fully} won the COCO instance segmentation challenge in 2016. 
It leverages the position sensitive score maps used in
the mask proposal mechanism of  \cite{Dai2016isfcn}.
At each location, it predicts a mask, as well as a category likelihood.
DeepMask \cite{Pinheiro2015} and SharpMask \cite{Pinheiro2016}
use a similar approach,
without the position-sensitive score maps.
Note that
most of these sliding window methods used an image pyramid to handle objects of
multiple sizes. Recently, \cite{Lin2017,Hu2016} proposed to use a feature
pyramid network instead of recomputing features on an image pyramid.

\eat{
does not use
box proposals, but instead predicts a per-pixel heatmap of detection
scores for each part of each object class (this is like semantic
segmentation, but extended by partitioning each class label into a
fixed number of parts).  These part heat maps are then combined, and
non-max suppression (NMS) is used to find high-scoring objects.
}

\begin{figure}
\begin{centering}
\includegraphics[scale=0.255]{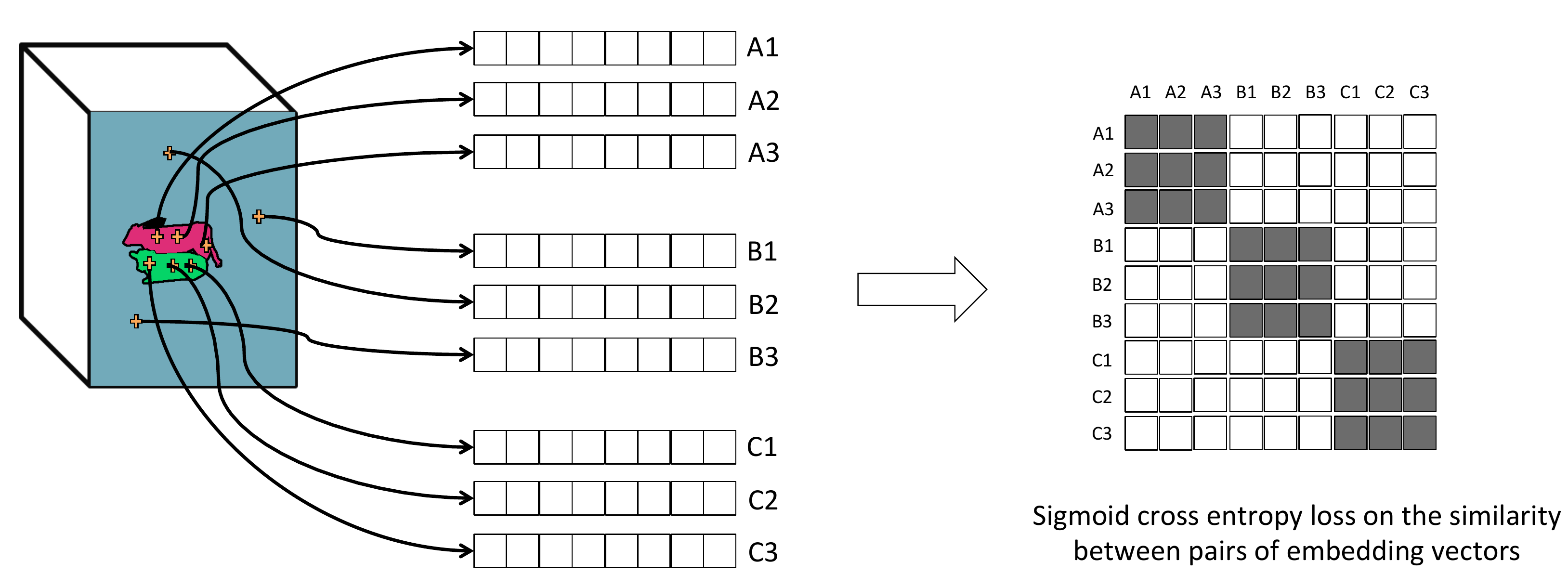}
\par
\end{centering}
\caption{
We compute the embedding loss by sampling the embedding vector of $K$ pixels within each instance, resulting in a total of $N K$ embedding vectors where $N$ is the number of instances in image.  
We compute the similarity between every pair of embedding vectors
$\sigma(e_p,e_q)$ as described in Eq. \ref{eq_s}. We want to learn a metric that returns a similarity of $1$ for a pair of embedding vectors that belong to the same instance, and $0$ for a pair of embedding vectors that belong to different instances. Thus, we add a cross entropy loss on the similarity between pairs by setting the ground-truth values to $0$ and $1$ based on whether embedding vectors belong to the same instance or not. 
\label{fig:EmbeddingLoss}
}
\end{figure}

The above methods all operate in parallel across the whole image.
Alternatively, we can run sequentially, extracting one object instance
at a time. \cite{Romera-Paredes2016} and \cite{Ren2017} are examples
of this approach. In particular, they extract features using a CNN and
then use an RNN to ``emit'' a binary mask (corresponding to an object
instance) at each step. The hidden state of the RNN is image-shaped
and keeps track of which locations in the image have already been
segmented. The main drawback of this approach is that it is
slow. Also, it is troublesome to scale to large images, since the
hidden state sequence can consume a lot of memory.

An another way to derive a variable number of objects (regions) is to
use the watershed algorithm. The method of \cite{Bai2016} predicts a
(discretized) energy value for each pixel. This energy surface is then
partitioned into object instances using the watershed algorithm. The
process is repeated independently for each class. The input to the
network is a 2d vector per pixel, representing the direction away from
the pixel's nearest boundary; these vectors are predicted given an RGB
input image. The overall approach shows state of the art results on
the CityScapes dataset. (A similar approach was used in
\cite{Uhrig2016}, except they used template matching to find the
instances, rather than watershed; also, they relied on depth data
during training.)

Kirrilov et al. \cite{Kirillov2016} also use the watershed algorithm to find candidate
regions, but they apply it to an instance-aware edge boundary map
rather than an energy function. They extract about 3000 instances
(superpixels), which become candidate objects. They then group (and
apply semantic labels) to each such region by solving a Multi-Cut
integer linear programming optimization problem.

Although the watershed algorithm has some appeal, these techniques
cannot group disconnected regions into a single instance (e.g., if the
object is partitioned into two pieces by an occluder,
such as the horse in Figure\ref{fig:ClassificationHeatmap}
who is occluded by its rider).
Therefore, we consider more general clustering
algorithms. Newel and Deng \cite{Newell2016} predict an objectness score for each
pixel, as well as a one-dimensional embedding for each pixel. They
first threshold on the objectness heatmap to produce a binary
mask. They then compute a 1d histogram of the embeddings of all the
pixels inside the mask, perform NMS to find modes, and then assign
each pixel in the mask to its closest centroid.

Our approach is related to \cite{Newell2016}, but differs in the
following important ways: (1) we use a different loss function when
learning the pixel similarity metric; (2) we use a different way of
creating masks, based on identifying the basins of attraction in
similarity space, rather than using a greedy clustering method; (3) we
learn a $D$-dimensional embedding per pixel, instead of using 1d
embeddings. As a consequence, our results are much better: we get a
\mAPr\ (at an \iou\ of 0.5) of \topmap\ on PASCAL VOC 2012 validation,
whereas \cite{Newell2016} gets 35.1\%.

\section{Method}
\label{sec:method}

In the following sections, we describe our method in more detail.

\begin{figure*}[t]
\begin{centering}
\includegraphics[scale=0.243]{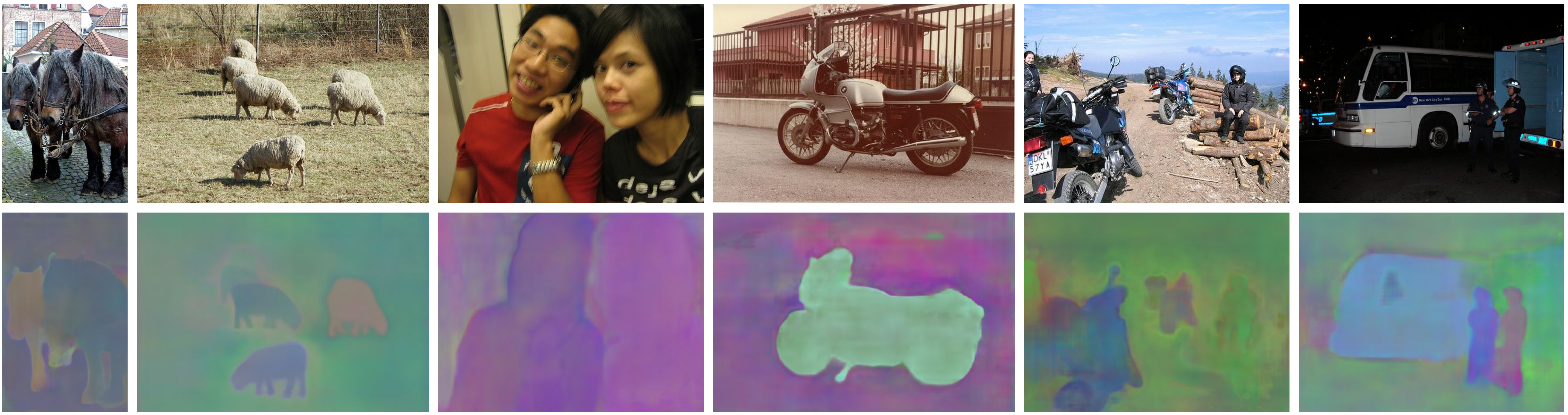}
\par
\end{centering}

\caption{Visualization of the embedding vectors by randomly projecting the
$64$ dimensional vectors into RGB space. The learned metric will move different instances
  of the same object category to different locations in embedding space.
  \label{fig:Visualization-of-embeddings}}
\end{figure*}

\subsection{Overview of our approach}

We start by taking a model that was pretrained for semantic
segmentation (see Section \ref{sec:deeplab} for details), and then
modify it to perform instance segmentation by adding two different
kinds of output ``heads''.
The first output head produces embedding vectors for each
pixel. Ideally, embedding vectors which are similar are more likely to
belong to the same object instance.    
In the second head, the model predicts a class label for the mask
centered at each pixel, as well as a confidence score that this pixel
would make a good ``seed'' for creating a mask. We sketch the overall
model in Figure~\ref{fig:Model}, and give the details below.

\subsection{Embedding model}
\label{sec:embedding}

We start by learning an embedding space, so that pixels that
correspond to the same object instance are close, and pixels that
correspond to different objects (including the background) are
far. The embedding head of our network takes as input a feature map
from a convolutional feature extractor (see Section
\ref{sec:deeplab}). It outputs a $[h, w, d]$ tensor (as shown in
Figure~\ref{fig:Model}), where $h$ is the image height, $w$ is the
image width and $d$ is the embedding space dimension (we use $64$
dimensional embeddings in our experiments). Thus each pixel $p$ in
image is represented by a $d$-dimensional embedding vector $e_p$.  

Given the embedding vectors, we can compute the similarity between pixels $p$ and $q$ as follows:
\begin{equation}
  \sigma(p,q) = \frac{2}{1 + \exp(||e_p - e_q||_2^2)}
  \label{eq_s}
\end{equation}
 We see that for pairs of pixels that are close in embedding space,
 we have $\sigma(p,q) = \frac{2}{1+e^0} = 1$,
 and for pairs of pixels that are far in embedding space,
 we have $\sigma(p,q) = \frac{2}{1+e^{\infty}} = 0$.

We train the network by minimizing the following loss:
\begin{align*}
\mathcal{L}_{e}
   &
  =-\frac{1}{|S|}\sum_{p,q \in S} w_{pq} \left[
    1_{\{y_{p}=y_{q}\}}\log(\sigma(p,q))    \right. \\
     & \left. + \ 1_{\{y_{p}\neq y_{q}\}}\log(1-\sigma(p,q)) \right]
\end{align*}
where $S$ is the set of pixels that we choose, $y_{p}$ is the instance
label of pixel $p$, and $w_{pq}$ is the weight of the loss on the similarity between $p$ and $q$. The
weights $w_{pq}$ are set to values inversely proportional to the size of
the instances $p$ and $q$ belong to, so the loss will not become
biased towards the larger examples. We normalize weights so that 
$\sum_{p,q}w_{pq}=1$.

During training,
we choose the set of pixels $S$ by randomly sampling $K$ points for
each object instance in the image.
For each pair of points, we compute the target label,
which is 1 if they are from the same instance,
and 0 otherwise,
as shown in Figure~\ref{fig:EmbeddingLoss}.
We then minimize
the cross-entropy loss for the $|S|^2$ set of points.
Our overall
procedure is closely related to the N-pairs loss used in
\cite{Sohn2016} for metric learning.

Figure~\ref{fig:Visualization-of-embeddings} illustrates the learned embedding for few example images. We randomly project the 64d vectors to RGB space and then visualize the resulting false-color image. We see that instances of the same class (e.g., the two people or the two motorbikes) get mapped to different parts of embedding space, as desired.

\subsection{Creating masks}
\label{sec:masks}

Once we have an embedding space, and hence a pairwise similarity
metric, we create a set of masks in the following way. We pick a
``seed'' pixel $p$, and then ``grow'' the seed by finding all the other
pixels $q$ that have a similarity with $p$ greater than a threshold
$\tau$: $m(p,\tau) = \{q: \sigma(p,q) \geq \tau\}$. Ideally, all the
pixels in the mask belong to the same object as the seed $p$. By
varying $\tau$, we can detect objects of different sizes.
In our experiments, we use $\tau \in \{0.25, 0.5, 0.75\}$.
(We also use a multi-scale representation of the image as input, as we
discuss in Section~\ref{sec:multiScale}.) 

We can implement this method efficiently as follows.
First
we compute a  tensor $A$
of size $[h,w,d]$ (where $h$ is the height of the image, $w$ is the
width, and $d$ is the emebdding dimension)
representing the embedding vector for  every pixel.
Next we compute
a second tensor $B$ of size $[k,d]$,
representing the embedding vector of the $K$ seed points.
We can compute the distance of each vector in $A$
to each vector in $B$ using
$A^2 + B^2 - 2 A \odot B$.
We can then select all the pixels that are sufficiently similar to each
of the seeds by thresholding this distance matrix.

\begin{figure}
\begin{centering}
\includegraphics[scale=0.27]{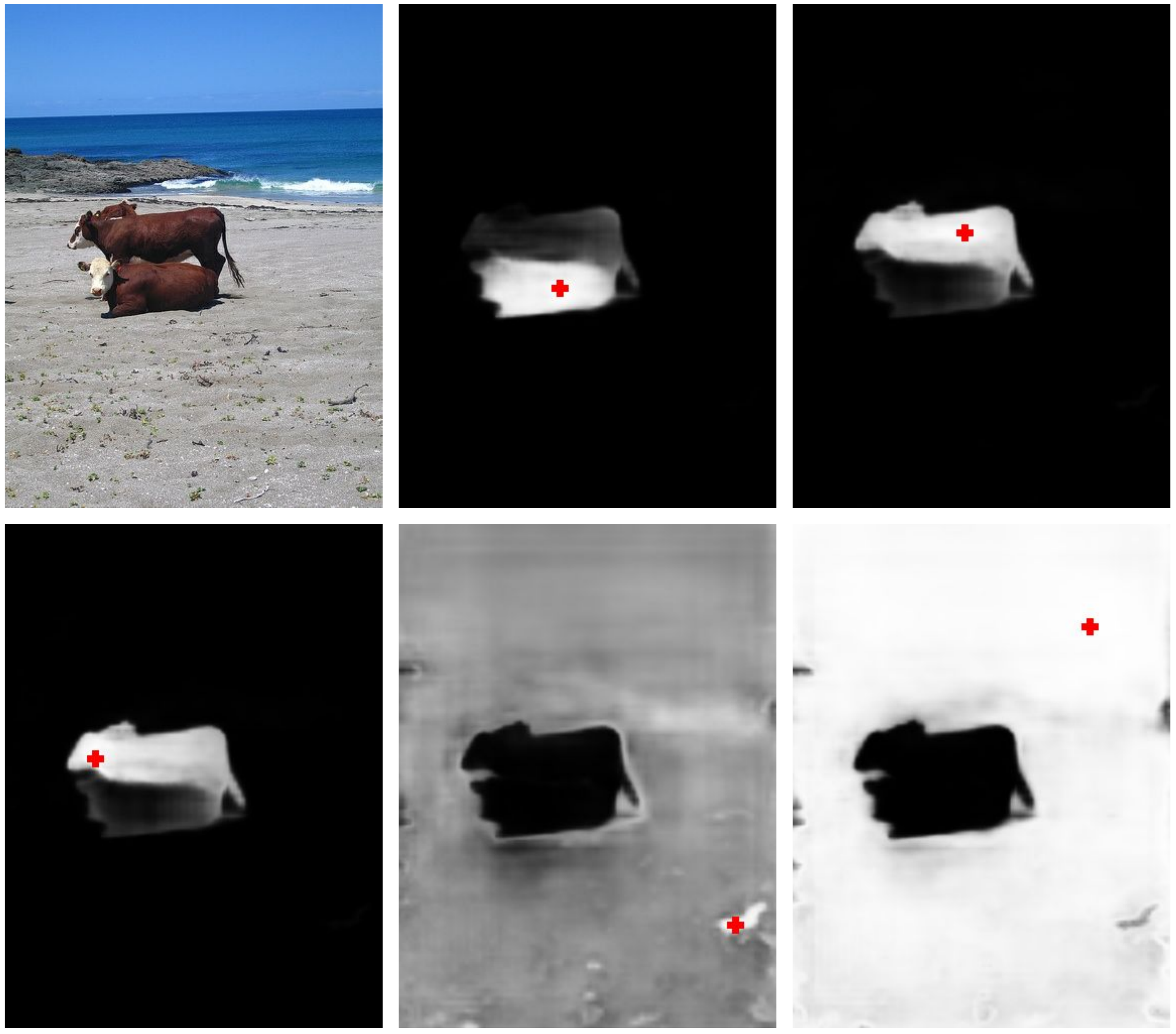}
\par
\end{centering}
\caption{
We visualize the similarity of each pixel and a randomly chosen seed pixel in each image. The randomly chosen seed pixel is shown by a red mark in the image. The brighter the pixels the higher the similarity. \label{fig:SoftMasks}
}
\end{figure}

Fig. \ref{fig:SoftMasks} shows examples of mask growing from randomly
picked seed pixel. The seed pixel is noted by a red mark in the
image. The similarity between every pixel and the seed pixel is
computed based on Eq. \ref{eq_s} and shown in each picture. One can
threshold the similarity values to generate a binary mask. 

 However, we still need a mechanism to choose the seeds. We propose to
 learn a ``seediness'' heatmap $S_p$, which tells us how likely
 it is that a mask grown from $p$  will be a good mask (one that overlaps with
 some ground truth mask by more than an \IOU\ threshold). We discuss
 how to compute the seediness scores 
 in Section~\ref{sec:seediness}.

\begin{figure*}[t]
\begin{centering}
\includegraphics[scale=0.275]{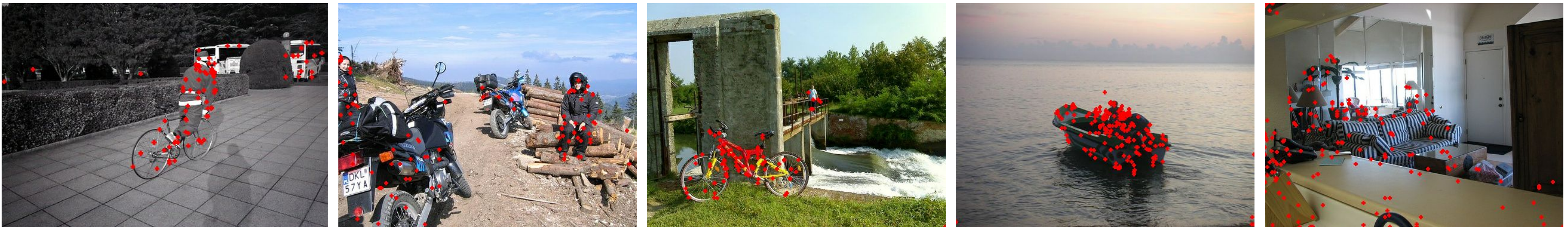}
\par
\end{centering}

\caption{Visualization of sampled seed pixels. Our method leverages the learned distance metric and masks classification scores to sample high-quality seed pixels that result in instance segments that have high recall and precision. \label{fig:Visualization-of-seed-pixels}
}
\end{figure*}

Once we have the seediness heatmap, we can greedily pick seed points
based on its modes (peaks). However, we also want to encourage spatial
diversity in the locations of our seed points, so that we ensure our
proposals have high coverage of the right set of masks. For this
reason, we also compute the distance (in embedding space) between each
point and all previously selected points and choose one that is far
from the already picked points (similar to the heuristic used by
Kmeans++ initialization \cite{Arthur2007}). More precisely, at step
$t$ of our algorithm, we select a seed point as 
follows:
\[
p_t = \arg \max_{p \not \in p_{1:t-1}} \left[
  \log(S_p) + \alpha \log(D(p, p_{1:t-1}) \right]
\]
where
\[
D(p, p_{1:t-1}) = \min_{q \in p_{1:t-1}} ||e_p - e_q||_2^2
\]

Selecting seed pixels with high seediness score
guarantees high precision, and selecting diverse seed points
guarantees high recall. Note that our sampling strategy is different
from the non-maximum suppression algorithm. In NMS, points that are
close in x-y image coordinate space are suppressed, while 
in our algorithm, we encourage diversity in the embedding space.

Once we have selected a seed, we select the best threshold $\tau$,
and then we can convert it into a mask,
$m_t = m(p_t, \tau)$, as we described above. Finally, we attach a confidence
score $s_t$ and a class label $c_t$ to the mask. To do this,  we
leverage a semantic-segmentation model which associates a predicted
class label with every pixel, as we explain in
Section~\ref{sec:class}.

\subsection{Classification and seediness model}
\label{sec:class}
\label{sec:seediness}

\begin{figure*}[t]
\begin{centering}
\includegraphics[scale=0.18]{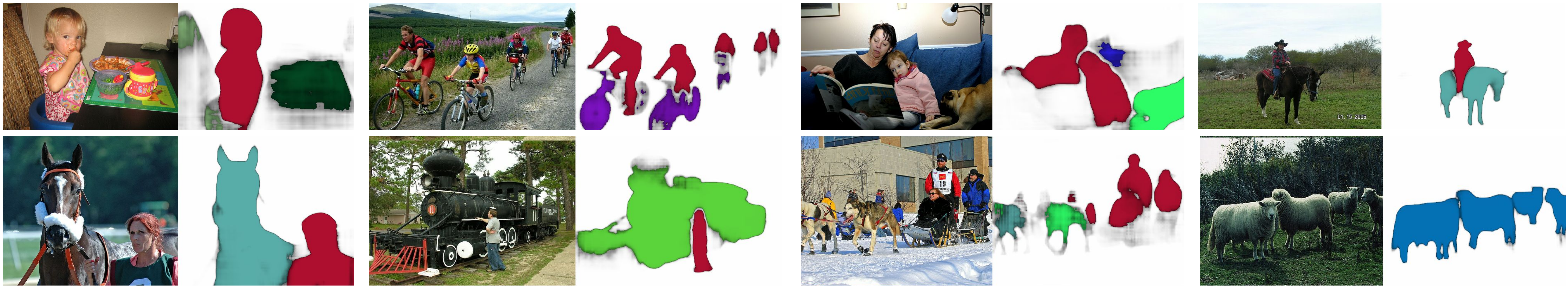}
\par
\end{centering}

\caption{Visualization of mask classification scores. 
The color at each pixel identifies the label of the mask that will be
chosen if that pixel is chosen as a seed point.
The pixels that are more likely to generate background masks are colored white. 
Darker colors correspond to pixels that will generate poor quality foreground masks.
Brighter colors correspond to pixels that generate high quality masks.
\label{fig:ClassificationHeatmap}
}

\end{figure*}

The mask classification head of our network takes as input a feature
map from a convolutional feature extractor (see Section
\ref{sec:deeplab}) and outputs a $[h, w, C+1]$ tensor as shown in
Fig. \ref{fig:Model}, where $C$ is the number of classes, and label 0
represents the background. In contrast to semantic segmentation, where
the pixels themselves are classified, here we classify the mask that
each pixel will generate if chosen as a seed. For example, assume a
pixel falls inside an instance of a horse. Semantic segmentation will
produce a high score for class ``horse'' at that pixel. However, our
method might instead predict background, if the given pixel is not a
good seed point for generating a horse mask.
We show examples of mask
classification heatmaps in Fig. \ref{fig:ClassificationHeatmap}.
We see that most pixels inside the object make good seeds,
but pixels near the boundaries (e.g., on the legs of the cows)
are not so good.

We train the model to emulate this behavior as follows. For each
image, we select $K=10$ random points per object instance and grow a mask
around each one. Let $m(p,\tau)$ be the mask generated from pixel
$p$, for a given similarity threshold $\tau$. If the proposed mask
overlaps with one of the ground truth masks by more than some fixed
\IOU\ threshold, we consider this a ``good'' proposal; we then copy
the label from the ground truth mask and assign it 
to pixel $p$. If the generated mask does not sufficiently overlap with
any of the ground truth masks, we consider this a ``bad'' proposal and
assign the background label to pixel $p$. We then convert the assigned
labels to one-hot form, and train using softmax cross-entropy loss,
for each of the $K$ chosen points per object instance. 
The classification model is fully convolutional, but we only evaluate
the loss at $N K$ points, where $N$ is the number of instances. Thus
our overall loss function has the form 
\[
\mathcal{L}_{cls}=
-\frac{1}{|S|} \sum_{p \in S}  \sum_{c=0}^C y_{pc} \log \mathcal{C}_{pc}
\]
where $\mathcal{C}_{pc}$ is the probability that the mask generated
from seed pixel $p$ belongs to class $c$.

To handle objects of
different sizes, we train a different classification model for each
value of $\tau$; in our experiments, we use
$\mathcal{T} = \{0.25, 0.5, 0.75, 0.9\}$.
Specifically, let $\mathcal{C}_{pc}^{\tau}$
represent the probability that pixel $p$ is a good seed for an
instance class $c$ when using similarity threshold $\tau$.

We now define the ``seediness'' of pixel $p$
to be
\[
S_p = \max_{\tau \in \mathcal{T}} \max_{c=1}^C \mathcal{C}_{pc}^{\tau}
\]
(Note that the max is computed over the object classes and not on the
background class.) Thus we see that the seediness tensor is computed
from the classification tensor. 

To understand why this is reasonable, suppose the background score at
pixel $p$ is very high, say 0.95. Then the max over foreground classes
is going to be smaller than 0.05 (due to the sum-to-one constraint on
$C_{pc}$), which means this is not a valid pixel to generate a
mask. But if the max value is, say, 0.6, this means this is a good
seed pixel, since it will grow into an instance of foreground class
with probability 0.6. 

Once we have chosen the best seed according to $S_p$,
we can find the corresponding best threshold $\tau$,
and label $c$ 
by computing
\[
(\tau_p,c_p) = \arg \max_{\tau \in \mathcal{T}, c \in 1:C}
 \mathcal{C}_{pc}^{\tau}
 \]
 The corresponding confidence score is given by
 \[
 s_p = \mathcal{C}_{p,c_p}^{\tau_p}
 \]
 
\subsection{Shared full image convolutional features}
\label{sec:deeplab}

Our base feature extractor is based on the DeepLab v2 model
\cite{Chen2017},
which in turn is based on resnet-101
\cite{He2016}.
We pre-train this for semantic segmentation on COCO,
as is standard practice for methods that compete on PASCAL VOC,
and then ``chop off'' the final layer. The Deeplab v2 model is fully
convolutional and runs on an image of size $[2h, 2w, 3]$ outputs a
[$\frac{h}{4}, \frac{w}{4}, 2048$] sized feature map which is the
input to both the embedding model and the classification/seediness
model. 

We can jointly train both output ``heads'',
and backprop into the shared ``body'',
by defining the loss
\[
\mathcal{L}=\mathcal{L}_{e}+\lambda\mathcal{L}_{cls}
\]
where $\mathcal{L}_{e}$ is the embedding loss, $\mathcal{L}_{cls}$ is
the classification loss, and $\lambda$ is a balancing coefficient. In
our experiments, we initially set $\lambda$ to $0$, to give the
embedding model time to learn proper embeddings. Once we have a
reasonable similarity metric, we can start to learn the mask
classification loss. We gradually increase $\lambda$ to a maximum
value of 0.2 (this was chosen empirically based on the validation
set). Both losses get back-propagated into the same feature
extractor. 
However, since we train on Pascal VOC, which is a small dataset, we
set the learning rate of the shared features to be smaller than for
the two output heads, so the original features (learned on COCO) do not change much. 

\subsection{Handling scale}
\label{sec:multiScale}

To handle objects of multiple scales,
we compute an image pyramid at 4 scales (0.25, 0.5, 1.0, 2.0),
and then run it through
our feature extractor 
These feature maps are then rescaled to the same size,
and averaged.
Finally, 
the result is fed into the two different heads,
as explained above.
(In the future,  we would like to investigate more efficient methods,
such as those discussed in \cite{Hu2016,Lin2017}.
that avoid having to run the base model 4 times.)

\section{Results}

  \begin{figure*}[!htbp]
\begin{centering}
\includegraphics[scale=0.27]{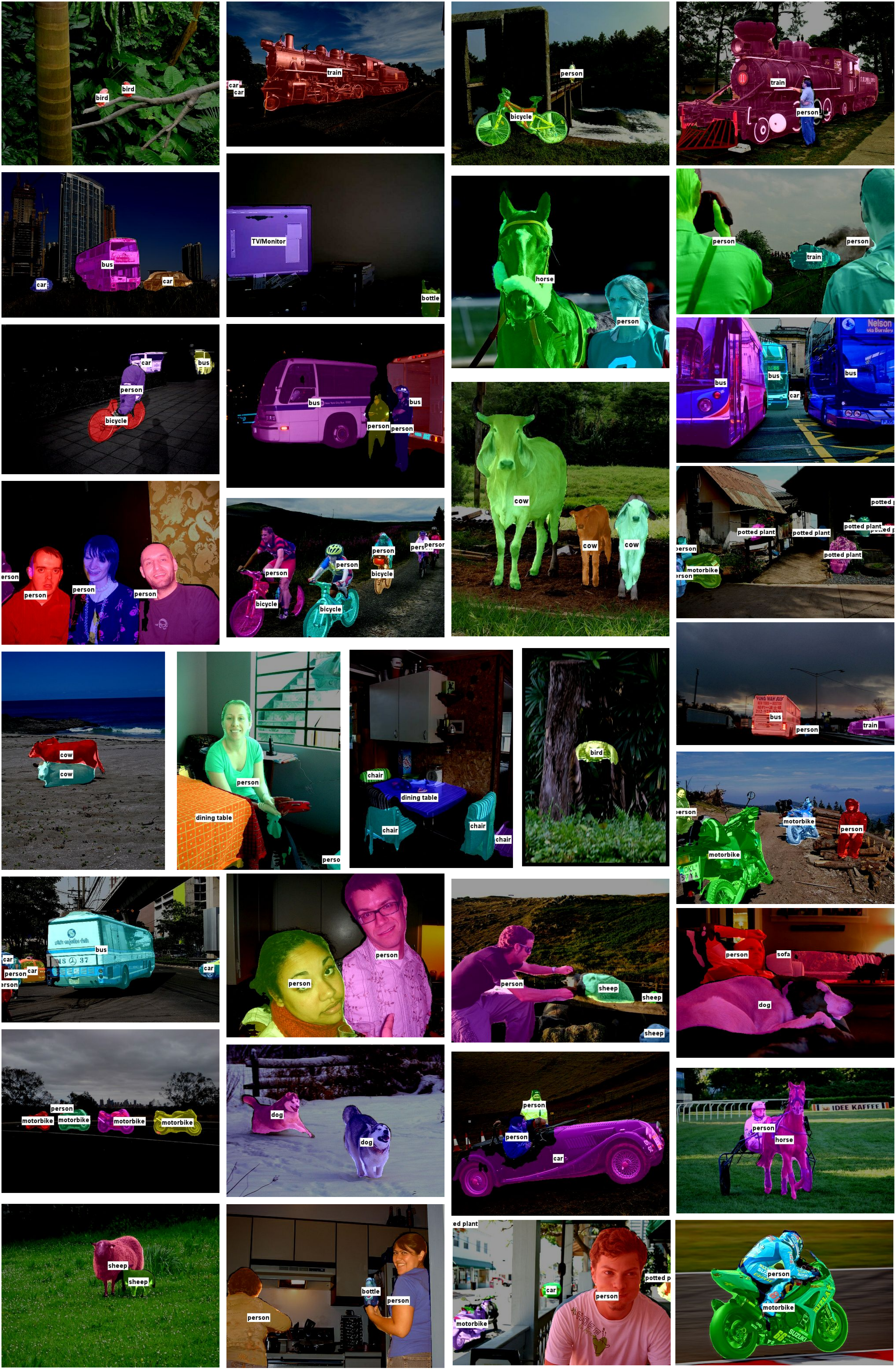}
\par
\end{centering}

\caption{Example instance segmentation results from our method.
\label{fig:SegmentOverlay}
}
\end{figure*}

\setlength{\tabcolsep}{0.25em} 
\begin{table*}[!htbp]
\scriptsize
  \caption{Per-class instance-level segmentation comparison using APr metric over 20 classes at 0.5, 0.6 and 0.7 IoU on the PASCAL VOC 2012 validation set. All numbers are in \%.}
\label{tab:results_iou}
\begin{center}
\begin{tabular}{| c | c | c c c c c c c c c c c c c c c c c c c c | c |}
\hline \normalsize{IoU score} & \normalsize{Method} & \begin{sideways}\normalsize{plane}\end{sideways} & \begin{sideways}\normalsize{bike}\end{sideways} & \begin{sideways}\normalsize{bird}\end{sideways} & \begin{sideways}\normalsize{boat}\end{sideways} & \begin{sideways}\normalsize{bottle}\end{sideways} & \begin{sideways}\normalsize{bus}\end{sideways} & \begin{sideways}\normalsize{car}\end{sideways} & \begin{sideways}\normalsize{cat}\end{sideways} & \begin{sideways}\normalsize{chair}\end{sideways} & \begin{sideways}\normalsize{cow}\end{sideways} & \begin{sideways}\normalsize{table}\end{sideways} & \begin{sideways}\normalsize{dog}\end{sideways} & \begin{sideways}\normalsize{horse}\end{sideways} & \begin{sideways}\normalsize{motor}\end{sideways} & \begin{sideways}\normalsize{person}\end{sideways} & \begin{sideways}\normalsize{plant}\end{sideways} & \begin{sideways}\normalsize{sheep}\end{sideways} & \begin{sideways}\normalsize{sofa}\end{sideways} & \begin{sideways}\normalsize{train}\end{sideways} & \begin{sideways}\normalsize{tv}\end{sideways}  & \normalsize{average} \\ \hline \hline
 
    & SDS \cite{Hariharan2014}         & 58.8 & 0.5  & 60.1 & 34.4 & 29.5 & 60.6 & 40.0 & 73.6 & 6.5  & 52.4 & 31.7 & 62.0 & 49.1 & 45.6 & 47.9 & 22.6 & 43.5 & 26.9 & 66.2 & 66.1 & 43.8 \\ 
\normalsize{0.5}
    & Chen et al. \cite{chen2015multi} & 63.6 & 0.3  & 61.5 & 43.9 & 33.8 & 67.3 & 46.9 & 74.4 & 8.6  & 52.3 & 31.3 & 63.5 & 48.8 & 47.9 & 48.3 & 26.3 & 40.1 & 33.5 & 66.7 & 67.8 & 46.3 \\ 
    & PFN \cite{liang2015proposal}     & 76.4 & 15.6 & 74.2 & 54.1 & 26.3 & 73.8 & 31.4 & 92.1 & 17.4 & 73.7 & 48.1 & 82.2 & 81.7 & 72.0 & 48.4 & 23.7 & 57.7 & 64.4 & 88.9 & 72.3 & 58.7 \\ 
    & MNC \cite{Dai2016mnc}            & -    & -    & -    & -    & -    & -    & -    & -    & -    & -    & -    & -    & -    & -    & -    & -    & -    & -    & -    & -    & 63.5 \\ 
    & Li et al. \cite{li2016fully}     & -    & -    & -    & -    & -    & -    & -    & -    & -    & -    & -    & -    & -    & -    & -    & -    & -    & -    & -    & -    & 65.7 \\ 
    & R2-IOS \cite{Liang2016}          & 87.0 & 6.1  & 90.3 & 67.9 & 48.4 & 86.2 & 68.3 & 90.3 & 24.5 & 84.2 & 29.6 & 91.0 & 71.2 & 79.9 & 60.4 & 42.4 & 67.4 & 61.7 & 94.3 & 82.1 & 66.7 \\
    & Assoc. Embedding \cite{Newell2016} & -    & -    & -    & -    & -    & -    & -    & -    & -    & -    & -    & -    & -    & -    & -    & -    & -    & -    & -    & -    & 35.1 \\ 
    & Ours                             & 69.7    & 1.2    & 78.2    & 53.8    & 42.2    & 80.1    & 57.4    & 88.8    & 16.0    & 73.2    & 57.9    & 88.4    & 78.9    & 80.0    & 68.0    & 28.0    & 61.5    & 61.3    & 87.5    & 70.4    & 62.1    \\ \hline

    & SDS \cite{Hariharan2014}         & 43.6 & 0    & 52.8 & 19.5 & 25.7 & 53.2 & 33.1 & 58.1 & 3.7  & 43.8 & 29.8 & 43.5 & 30.7 & 29.3 & 31.8 & 17.5 & 31.4 & 21.2 & 57.7 & 62.7 & 34.5 \\
\normalsize{0.6} 
    & Chen et al. \cite{chen2015multi} & 57.1 & 0.1  & 52.7 & 24.9 & 27.8 & 62.0 & 36.0 & 66.8 & 6.4  & 45.5 & 23.3 & 55.3 & 33.8 & 35.8 & 35.6 & 20.1 & 35.2 & 28.3 & 59.0 & 57.6 & 38.2 \\
    & PFN \cite{liang2015proposal}     & 73.2 & 11.0 & 70.9 & 41.3 & 22.2 & 66.7 & 26.0 & 83.4 & 10.7 & 65.0 & 42.4 & 78.0 & 69.2 & 72.0 & 38.0 & 19.0 & 46.0 & 51.8 & 77.9 & 61.4 & 51.3 \\
    & R2-IOS \cite{Liang2016}          & 79.7 & 1.5  & 85.5 & 53.3 & 45.6 & 81.1 & 62.4 & 83.1 & 12.1 & 75.7 & 20.2 & 81.5 & 49.7 & 63.9 & 51.2 & 35.7 & 56.2 & 56.7 & 87.9 & 78.8 & 58.1 \\
    & Ours                             & 64.2    & 0.1    & 64.8    & 37.2    & 34.5    & 73.5    & 50.6    & 84.7    & 8.9    & 59.3    & 48.2    & 84.3    & 65.1    & 69.6    & 56.6    & 14.9    & 51.8    & 50.7    & 81.7    & 64.4    & 53.3    \\ \hline

    & SDS \cite{Hariharan2014}         & 17.8 & 0    & 32.5 & 7.2  & 19.2 & 47.7 & 22.8 & 42.3 & 1.7  & 18.9 & 16.9 & 20.6 & 14.4 & 12.0 & 15.7 & 5.0  & 23.7 & 15.2 & 40.5 & 51.4 & 21.3 \\
\normalsize{0.7}
    & Chen et al. \cite{chen2015multi} & 40.8 & 0.07 & 40.1 & 16.2 & 19.6 & 56.2 & 26.5 & 46.1 & 2.6  & 25.2 & 16.4 & 36.0 & 22.1 & 20.0 & 22.6 & 7.7  & 27.5 & 19.5 & 47.7 & 46.7 & 27.0 \\
    & PFN \cite{liang2015proposal}     & 68.5 & 5.6  & 60.4 & 34.8 & 14.9 & 61.4 & 19.2 & 78.6 & 4.2  & 51.1 & 28.2 & 69.6 & 60.7 & 60.5 & 26.5 & 9.8  & 35.1 & 43.9 & 71.2 & 45.6 & 42.5 \\
    & MNC \cite{Dai2016mnc}            & -    & -    & -    & -    & -    & -    & -    & -    & -    & -    & -    & -    & -    & -    & -    & -    & -    & -    & -    & -    & 41.5 \\ 
    & Li et al. \cite{li2016fully}     & -    & -    & -    & -    & -    & -    & -    & -    & -    & -    & -    & -    & -    & -    & -    & -    & -    & -    & -    & -    & 52.1 \\ 
    & R2-IOS \cite{Liang2016}          & 54.5 & 0.3  & 73.2 & 34.3 & 38.4 & 71.1 & 54.0 & 76.9 & 6.0  & 63.3 & 13.1 & 67.0 & 26.9 & 39.2 & 33.2 & 25.4 & 44.8 & 45.4 & 81.5 & 74.6 & 46.2 \\ 
    & Assoc. Embedding \cite{Newell2016} & -    & -    & -    & -    & -    & -    & -    & -    & -    & -    & -    & -    & -    & -    & -    & -    & -    & -    & -    & -    & 26.0 \\ 
    & Ours                             & 53.0    & 0.0    & 51.8    & 24.9    & 21.9    & 69.2    & 40.1    & 76.6    & 4.1    & 43.1    & 21.1    & 74.4    & 44.7    & 54.3    & 40.3    & 7.5   & 40.5    & 39.6    & 69.5    & 52.6    & 41.5    \\ \hline
\end{tabular}
\end{center}
\end{table*}

In this section, we discuss our experimental results.

 \subsection{Experimental Setup}

We follow the experimental protocol that is used by previous
state of the art methods
\cite{Liang2016,li2016fully,liang2015proposal,chen2015multi,Hariharan2014}.
In particular,
we train on the PASCAL VOC 2012 training
set, with additional instance mask annotation from
\cite{Hariharan2011},
and we evaluate on the PASCAL VOC 2012
validation set.

After training the model,
we compute a precision-recall curve for each class,
based on all the test data.
This requires a definition of what we mean by a true and false positive. We follow standard practice
and say that a predicted mask 
that has an intersection over union \iou\ with a true mask above
some threshold $\iouthresh$ (e.g. $50\%$)
is a true positive, unless the true mask is already detected, in which case the
detection is a false positive.
We provide results for three \iou\ thresholds: $\{0.5, 0.6, 0.7\}$ similar to previous work.
We then compute the area under the PR curve,
known as the ``average precision''  or \APr\ score
\cite{Hariharan2011}.
Finally, we average this over classes, to get the mean average
precision or \mAPr\ score.

We can also evaluate the quality of our method as a ``class agnostic'' region proposal generator.
There are two main ways to measure quality in this case.
The first is to plot the recall (at a fixed \iou) vs the number of
proposals.
The second is to plot the recall vs the \iou\ threshold
for a fixed number of proposals; the area under this curve is known as
the ``average recall'' or AR  \cite{Hosang2015}.
\eat{
We compute a separate such curve (and AR summary statistic) for each
number of proposals; to be comparable
with other works (e.g., \cite{Pinheiro2016,Hu2016}),
we will use $N \in \{10, 100, 1000\}$.
(We can also plot the recall at a fixed IoU threshold vs the number of
proposals.)
}

\subsection{Preprocessing}

We use the following data augmentation components during training:

\textbf{Random Rotation}: We rotate the training images by a random degree in the range of $[-10, 10]$.
 
\textbf{Random Resize}: We resize the input image during the training phase with a random ratio in the range of $[0.7, 1.5]$.

\textbf{Random Crop}: We randomly crop the images during the training phase. At each step, we randomly crop $100$ windows. 
We randomly sample one image weighted towards cropped windows that have more object instances inside.

\textbf{Random Flip}: We randomly flip our training images horizontally.

\subsection{Results on Pascal VOC 2012}

\begin{table}
\begin{centering}
\begin{tabular}{|c||c|c|c|c|c|c|}
\hline 
$\alpha$ & 0.1 & 0.2 & 0.3 & 0.4 & 0.5 & 0.6\tabularnewline
\hline 
\hline 
$mAP^{r}$ & 61.4 & 61.7 & \textbf{62.1} & 61.6 & 61.6 & 61.5\tabularnewline
\hline 
\end{tabular}
\par\end{centering}

\caption{We find out that the best performing $\alpha$ parameter for sampling
seed points is $0.3$. In the table, we compare the mAPr performance
for different values of $\alpha$.
\label{tab:AlphaSweep}
}
\end{table}

We first tried different values of $\alpha$ (which trades off diversity with
seediness when picking the next seed) to find the best performing seed
sampling strategy. The results are shown for different values of
$\alpha$ in Table \ref{tab:AlphaSweep}.
We also tried
various sizes of embeddings from $2$ to $256$;
64 was best (on a validations set).
We furthermore analyze the performance of our model given different number of mask proposals (number of sampled seed points) in Table \ref{tab:NumSampleSweep}.
Our model reaches a $mAP^r$ performance of $59.7$ by only proposing $10$ regions.
In Table \ref{tab:NumSampleSweep}, we also show the class agnostic average recall for different number of mask proposals.

\begin{table}
\tiny
\begin{centering}
\begin{tabular}{|c|c|c|c|c|c|c|c|c|c|c|c|c|c|}
\hline 
Number of samples & 10 & 20 & 30 & 40 & 50 & 60 & 70 & 80 & 90 & 100 & 200 & 500 & 1000\tabularnewline
\hline 
\hline 
AR at $50\%$ IOU & 55.8 & 58.5 & 61.0 & 63.0 & 65.2 & 66.7 & 68.4 & 69.8 & 70.8 & 71.9 & 77.4 & 82.2 & 83.6\tabularnewline
\hline 
$mAP^{r}$ with $\alpha=0.2$ & 59.7 & 60.7 & 60.7 & 60.9 & 61.2 & 61.3 & 61.1 & 61.1 & 61.0 & 61.0 & 61.4 & 61.5 & 61.7\tabularnewline
\hline 
\end{tabular}
\par\end{centering}

\caption{We analyze the performance of our model given the number of sampled
  seed points.
\label{tab:NumSampleSweep}
}

\end{table}

Figure~\ref{fig:SegmentOverlay} shows some qualitative results,
and
Table~\ref{tab:results_iou} shows our quantitative results.
In terms of mAP performance,
we rank 4th at 0.5 \iou\ threshold,
2nd at 0.6 \iou\ threshold,
and tied 3rd at 0.7 \iou\ threshold.
So our method is competitive, if not state of the art.

Regarding performance for individual classes,
we see that we do very well on large objects,
such as trains, dogs, and motorbikes,
but we do very poorly in the bicycle category.
(See Fig. \ref{fig:ClassificationHeatmap} for some examples.)
This is also true of the other methods.
The reason is that there is a large discrepancy between
the quality of the segmentation masks
in the training set of \cite{Hariharan2011}
compared to the PASCAL test set.
In particular, the training set uses a coarse mask covering the entire
bicycle, whereas the test set segments out each spoke of the wheels
individually.

\section{Conclusion and future work}

We have proposed a novel approach to the semantic instance segmentation problem, and obtained promising preliminary results on the PASCAL VOC dataset. In the future, we would like to evaluate our method on COCO and Cityscapes. We would also like to devise a way to perform region growing in a differentiable way so that we can perform end-to-end training.

{\small
\bibliographystyle{ieee}
\bibliography{bib}
}

\end{document}